\documentclass[runningheads]{llncs}
\usepackage{graphicx}
\usepackage{cite}
\usepackage{tikz}
\usepackage{comment}
\usepackage{subfiles}
\usepackage{amsmath,amssymb} %
\usepackage{color}
\usepackage{booktabs} 
\usepackage{subcaption}

\usepackage[capitalize]{cleveref}
\crefname{section}{Sec.}{Secs.}
\Crefname{section}{Section}{Sections}
\Crefname{table}{Table}{Tables}
\crefname{table}{Tab.}{Tabs.}

\usepackage[accsupp]{axessibility}  %

\usepackage{xspace}

\makeatletter
\DeclareRobustCommand\onedot{\futurelet\@let@token\@onedot}
\def\@onedot{\ifx\@let@token.\else.\null\fi\xspace}

\def\eg{\emph{e.g}\onedot} 
\def\ie{\emph{i.e}\onedot}

\def\etal{\emph{et al}\onedot}
\makeatother

\begin{document}
\pagestyle{headings}
\mainmatter
\def\ECCVSubNumber{4748}  %

\title{Faster VoxelPose: Real-time 3D Human Pose Estimation by Orthographic Projection} %

\titlerunning{Faster VoxelPose}
\author{Hang Ye\inst{1*} \and
Wentao Zhu\inst{2,3*} \and
Chunyu Wang\inst{4+} \and \\
Rujie Wu\inst{2,3} \and
Yizhou Wang\inst{2,3,5}
}
\authorrunning{H. Ye et al.}
\def\thefootnote{*}\footnotetext{Equal contribution.}\def\thefootnote{\arabic{footnote}}
\def\thefootnote{+}\footnotetext{Corresponding author.}\def\thefootnote{\arabic{footnote}}
\institute{
Yuanpei College, Peking University \and
Center on Frontiers of Computing Studies, Peking University \and
School of Computer Science, Peking University \and
Microsoft Research Asia \and
Inst. for Artificial Intelligence, Peking University \\
\email{\{yehang, wtzhu, wu\_rujie, yizhou.wang\}@pku.edu.cn, chnuwa@microsoft.com}}
\maketitle

\begin{abstract}
While the voxel-based methods have achieved promising results for multi-person 3D pose estimation from multi-cameras, they suffer from heavy computation burdens, especially for large scenes.
We present \textit{Faster VoxelPose} to address the challenge by re-projecting the feature volume to the three two-dimensional coordinate planes and estimating $X, Y, Z$ coordinates from them separately. To that end, we first localize each person by a 3D bounding box by estimating a 2D box and its height based on the volume features projected to the xy-plane and z-axis, respectively. Then for each person, we estimate partial joint coordinates from the three coordinate planes separately which are then fused to obtain the final 3D pose. The method is free from costly 3D-CNNs and improves the speed of VoxelPose by ten times and meanwhile achieves competitive accuracy as the state-of-the-art methods, proving its potential in real-time applications. 

\end{abstract}

\section{Introduction}
\label{sec:intro}

Estimating 3D human pose from RGB images is a fundamental problem in computer vision. It not only paves the way for some important downstream tasks such as action recognition~\cite{wang2013approach,wang2016mining,wang2016recognizing,DBLP:conf/aaai/YanXL18, Shi2019-pq} and human-computer interaction~\cite{CoolMoves,10.1145/3173574.3173588}, but also enables a wide range of applications, \eg sports analysis~\cite{ramanathan_cvpr16, Bridgeman_2019_CVPR_Workshops} and virtual avatar animation~\cite{zhu_2020_eccv_nba, weng2019photo}.

While many works~\cite{wang2014robust,wang2018robust,ci2019optimizing, pavllo20193d, ma2021context, ZhouHanICCV19, zheng20213d} address monocular 3D pose estimation, their application in serious scenarios is limited because of the degraded accuracy~\cite{Li_2019_CVPR, Sharma_2019_ICCV}. In addition, monocular human pose estimation struggles when occlusion occurs which is ubiquitous in natural images~\cite{DBLP:conf/eccv/DongSZLZB20, Cheng2019-qq}. %
As a result, the state-of-the-art 3D human pose estimation results are usually obtained via multi-camera systems which consist of a group of synchronized and calibrated wide-baseline cameras~\cite{ voxelpose,zhang2022voxeltrack, zhang2020fusing,zhang2021adafuse, Wu_undated-wk, dong2019fast, Lin_2021_CVPR, Belagiannis_2014_CVPR}.

\begin{figure}[t]
  \centering
  \includegraphics[width=0.7\linewidth]{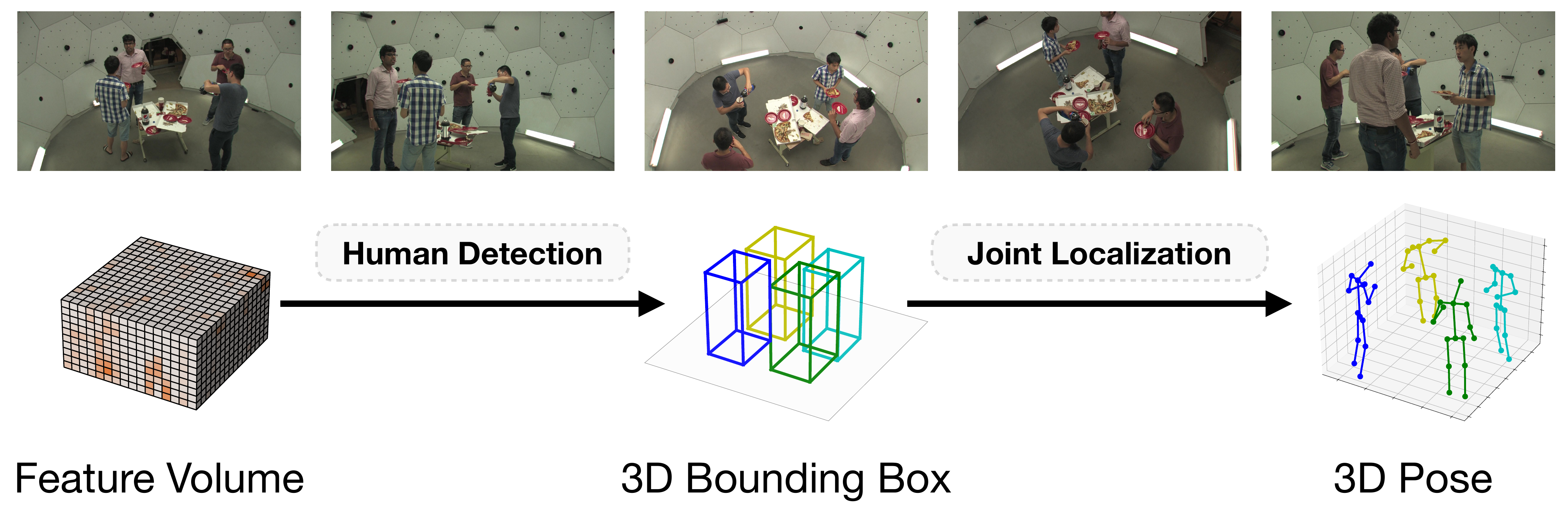}
  \caption{\textbf{Multi-view 3D Pose Estimation.} Given multi-view images and camera parameters, the task aims to estimate the 3D poses of all people in the world coordinates. Similar to \cite{voxelpose}, our approach is based on the volumetric representation and detects 3D box as an intermediate step.}
  \label{fig:teaser}
\end{figure}

Simple triangulation~\cite{10.5555/861369, HUMBI} can achieve accurate 3D pose estimates if the 2D poses in all views are accurate. However, 2D pose estimates may have errors in practice especially when occlusion occurs. To address the problem, voxel-based methods~\cite{voxelpose, VoxelTrack, DBLP:conf/iccv/QiuWWWZ19,Iskakov2019-fe, Wu_undated-wk, reddy2021tessetrack} have been proposed which inversely project 2D features or heatmaps in each view to the 3D space and then fuse the multi-view features. The resulting feature volume is more robust to occlusions in individual cameras. Then they apply a 3D-CNN to estimate the 3D positions of the body joints from the feature volume. 
While these methods achieve very accurate results, the computation complexity increases cubically with space size. As a result, they cannot support real-time inference for large scenes such as sports stadiums or retail stores.

In this work, we present Faster VoxelPose which is about ten times faster than VoxelPose on the common benchmarks and more importantly scales gracefully to large spaces. Inspired by technical drawing where a 3D object is often unambiguously represented by three 2D orthographic projections, \ie plan, elevation and section, we re-project the previously fused 3D volumetric features to the three 2D coordinate planes by orthographic projection and estimate partial coordinates, \eg $xy$, $xz$ and $yz$, of a 3D pose from each of the 2D planes, which are then fused by a tiny network to predict $xyz$. The main advantage of the method is that we can replace the expensive 3D-CNNs with 2D-CNNs which reduces the computation cost from $O(n^3)$ to $O(n^2)$ where $n$ is the spatial resolution. However, the factorization brings two new challenges. First, people that are far away in the 3D space could overlap in some planes after re-projection, which may bring severe ambiguity to the corresponding features. Second, the estimation results may be inconsistent across planes so we need a strategy to aggregate the contradictory predictions.

We address the challenges from two aspects. Firstly, as shown in Fig.~\ref{fig:teaser}, we present \textit{Human Detection Networks} (HDN) to estimate a \emph{tight} 3D box for each person which is used to filter out the features of other people. By contrast, VoxelPose~\cite{voxelpose} use a loose fixed-size 3D bounding box. In particular, we re-project the 3D feature volume to the $xy$ plane by max-pooling along the $z$ axis (bird's-eye view), and apply a 2D-CNN to localize people by a 2D box in the $xy$ plane. Then, for each bounding box, we obtain a 1D ``column'' feature representation from the volume at the box center along the $z$ axis, and apply a 1D-CNN to estimate the vertical position of the box center.

Then we present \textit{Joint Localization Networks} to estimate a 3D pose for each 3D box. We first mask out the features in the volume which are outside the box to reduce the impact of other people, obtaining \textit{person-specific} feature volume. We re-project the masked volume to the three coordinate planes and estimate the $X$, $Y$ and $Z$ coordinates, respectively. For each coordinate, we have two predictions from two planes. It is probable that the two predictions are contradictory so we propose a fusion network to learn a weight for each prediction and aggregate them to obtain the final 3D pose. 

Our approach achieves competing results as the baseline method which uses 3D-CNN. But ours is about $10$ times faster than it (speed improvement is larger for larger scenes). Our contributions are four-fold: 1) We design a lightweight framework for efficient training and inference of the multi-view multi-person 3D pose estimation problem. Our approach demonstrates that 3D human detection and pose estimation can be resolved on the re-projected 2D feature maps with careful design. 2) We propose a novel 3D human detector that disentangles ground plane localization and height estimation. 3) We utilize 3D bounding box for feature masking, which contributes to person-specific feature volume and improves joint localization accuracy. 4) We deploy the confidence regression networks to adaptively fuse the estimates on the re-projected planes to compensate for their individual accuracy loss. While we focus on pose estimation in this work, the idea may also benefit other voxel-based tasks such as object detection\cite{liu-voxel21, rukhovich2022imvoxelnet, DBLP:journals/corr/abs-2107-02980} and shape completion\cite{wang2021voxel}.

\section{Related Work}
\label{sec:related}

\subsection{Multi-view 3D Pose Estimation}

For the single-person case, the key is to handle 2D pose estimation errors in individual planes. Iskakov \etal~\cite{Iskakov2019-fe} designed differentiable triangulation which uses joint detection confidence in each camera view to learn the optimal triangulation weights. Pavlakos \etal~\cite{pavlakos17harvesting} applied CNN with 3D PSM for markerless motion capture. Qiu \etal~\cite{DBLP:conf/iccv/QiuWWWZ19} used epipolar lines to guide cross-view feature fusion followed by a recurrent PSM. Epipolar transformer~\cite{he2020epipolar} extended~\cite{DBLP:conf/iccv/QiuWWWZ19} to handle dynamic cameras. Generally speaking, single-person 3D pose estimation has achieved satisfactory results when there are sufficient cameras to guarantee that every body joint can be seen from at least two cameras.

Multi-person 3D human pose estimation is more challenging because it needs to solve two additional sub-tasks: 1) Identifying joint-to-person association in different views. 2) Handling mutual occlusions among the crowd. To address the first challenge, various association strategies are proposed based on re-id features~\cite{dong2019fast}, dynamic matching~\cite{Belagiannis_2014_CVPR}, 4D graph cut~\cite{20204DAssociation}, and plane sweep stereo~\cite{Lin_2021_CVPR}. However, in crowded scenes, noisy 2D pose estimates would harm their accuracy. To address the second challenge, Belagiannis \etal~\cite{Belagiannis_2014_CVPR} extended PSM for multi-person. Wang \etal~\cite{wang2021mvp} propose a transformer-based direct regression model with projective attention.

Recently, voxel-based methods~\cite{voxelpose, VoxelTrack, Wu_undated-wk, Narapureddy-2021-126995} are proposed to avoid making decisions in each camera view. Instead, they fuse multi-view features in the 3D space and only make the decision there. Such methods are free from pairwise reasoning of camera views and enable learning human posture knowledge in a data-driven way. However, the computation-intensive 3D convolutions prevent these approaches from being real-time and applicable to large spaces. Our method enjoys the benefit of volumetric feature aggregation, meanwhile being significantly faster and more scalable.

\subsection{Efficient Human Pose Estimation}

Designing efficient human pose estimators has been intensively studied for practical usage. For extracting 2D pose from images, state-of-the-art methods~\cite{Zhang_2019_CVPR, Xu_2021_CVPR, DBLP:journals/corr/abs-2108-02092, DBLP:conf/ijcai/ShenYNMGLRW21, VNect_SIGGRAPH2017} have achieved real-time inference speed. In terms of multiview 3D pose estimation, Bultman \etal~\cite{DBLP:journals/corr/abs-2106-14729} explores an efficient system using edge sensors. Remelli \etal~\cite{Remelli_2020_CVPR} and Fabbri \etal~\cite{Fabbri_2020_CVPR} adopt encoder-decoder networks to reduce computation, but they are not applicable to the multi-person setting. Most recently, Lin \etal~\cite{Lin_2021_CVPR} and Wang \etal~\cite{wang2021mvp} present alternative solutions to volumetric methods~\cite{voxelpose, VoxelTrack, Wu_undated-wk} and show some speed improvement. Nevertheless, these methods are capped in terms of scalability, which prevents them from being deployed to large scenes. Our method is complementary to state-of-the-art lightweight 2D pose estimators, and can further improve the speed of other volumetric methods~\cite{Fabbri_2020_CVPR, Wu_undated-wk, Narapureddy-2021-126995}. 

\section{Method}
\label{sec:method}

\subsection{Overview}
\begin{figure*}[t]
  \centering
  \includegraphics[width=\linewidth]{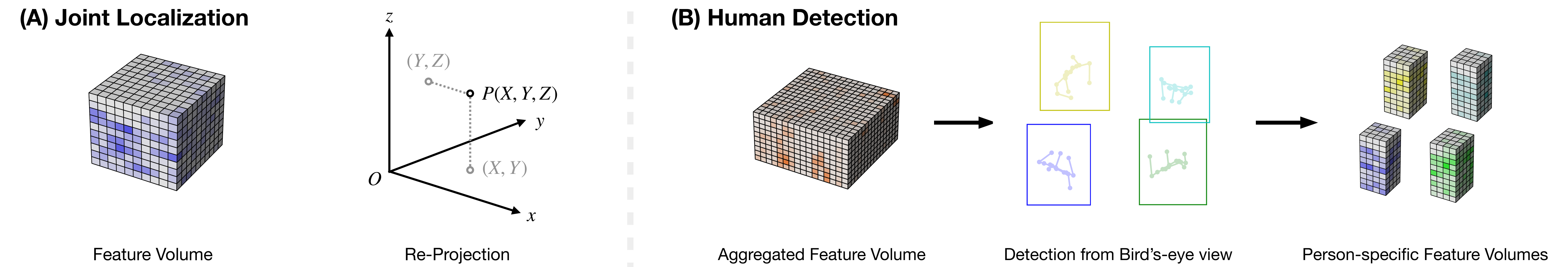}
  \caption{\textbf{Problem Decomposition.} (A): Considering a single person, we re-project its feature volume to the coordinate planes with orthographic projection. The partial coordinates can be estimated by 2D CNN and assembled to 3D estimation. (B): Multi-person brings the extra challenge of ambiguity and occlusion. Nonetheless, people can be easily isolated from the bird's-eye view of the aggregated feature volume. Based on the intuitive ideas, we develop the lightweight \textit{Joint Localization Networks} and \textit{Human Detection Networks} respectively. }
  \label{fig:int}
\end{figure*}

Without loss of generality, we explain our motivation with a simple case in which there is only one person. As shown in Fig.~\ref{fig:int} (A), the input to our approach is a 3D feature volume $\mathbf{V} \in \mathbb{R}^{K \times L \times W \times H}$ which is constructed by back-projecting the 2D pose heatmaps in multiple cameras to the 3D voxel space \cite{voxelpose}. The 2D pose heatmaps are extracted from the images using an off-the-shelf pose estimation model ~\cite{sun2019deep}. $L \times W \times H$ represents the number of voxels that are used to discretize the space and $K$ represents the number of joint types. The volume approximately encodes the per-voxel likelihood of body joints. 

In Fig.~\ref{fig:int} (A), we show a 3D joint of interest, \eg a shoulder joint, as $P=(X, Y, Z)$. In general, the corresponding feature volume should have a distinctive pattern around $P$ so that it can be localized by expensive 3D-CNNs~\cite{voxelpose}. To reduce the computation cost, we re-project the volume to the three coordinate planes (\ie the $xy$, $yz$, $xz$ planes), respectively, resulting in three 2D feature maps. We can imagine that there are also distinctive patterns at the corresponding locations of each 2D feature map, \eg $(X, Y)$ at the $xy$ plane, which can be similarly detected by 2D-CNNs. Then the 3D position of $P$ can be assembled from the estimated coordinates in the three planes. 

However, when we apply the idea to the multi-person scenario, we are confronted with new challenges. The features of different people may be mixed together after being projected to the coordinate planes even when they are far away from each other in the 3D space. This may corrupt the pose estimation accuracy. Inspired by top-down 2D pose estimation~\cite{fang2017rmpe}, the problem can be alleviated by ``cropping'' the person from the overall 3D space and only projecting features belonging to the person to the planes. So the remaining task is to detect each person in the 3D space efficiently. We utilize the prior that people barely overlap along the $z$ axis, therefore they can be easily detected in the bird's-eye view as shown in Fig.~\ref{fig:int} (B).

We take a two-phase approach to address the challenges. In the first phase, we present \textit{Human Detection Networks} (Section \ref{sec:hdn}) which efficiently detects all people from the bird's-eye view by 3D bounding boxes, ensuring that only the person-of-interest features are passed to the next phase.  The second phase conducts fine-grained pose estimation for each person with \textit{Joint Localization Networks} (Section \ref{sec:JLN}), which is greatly eased since occlusion and distraction are mostly eliminated in the first phase. Importantly, all the operators in the networks are on 2D and 1D features, which boosts the speed.

\begin{figure*}[t]
  \centering
  \includegraphics[width=\linewidth]{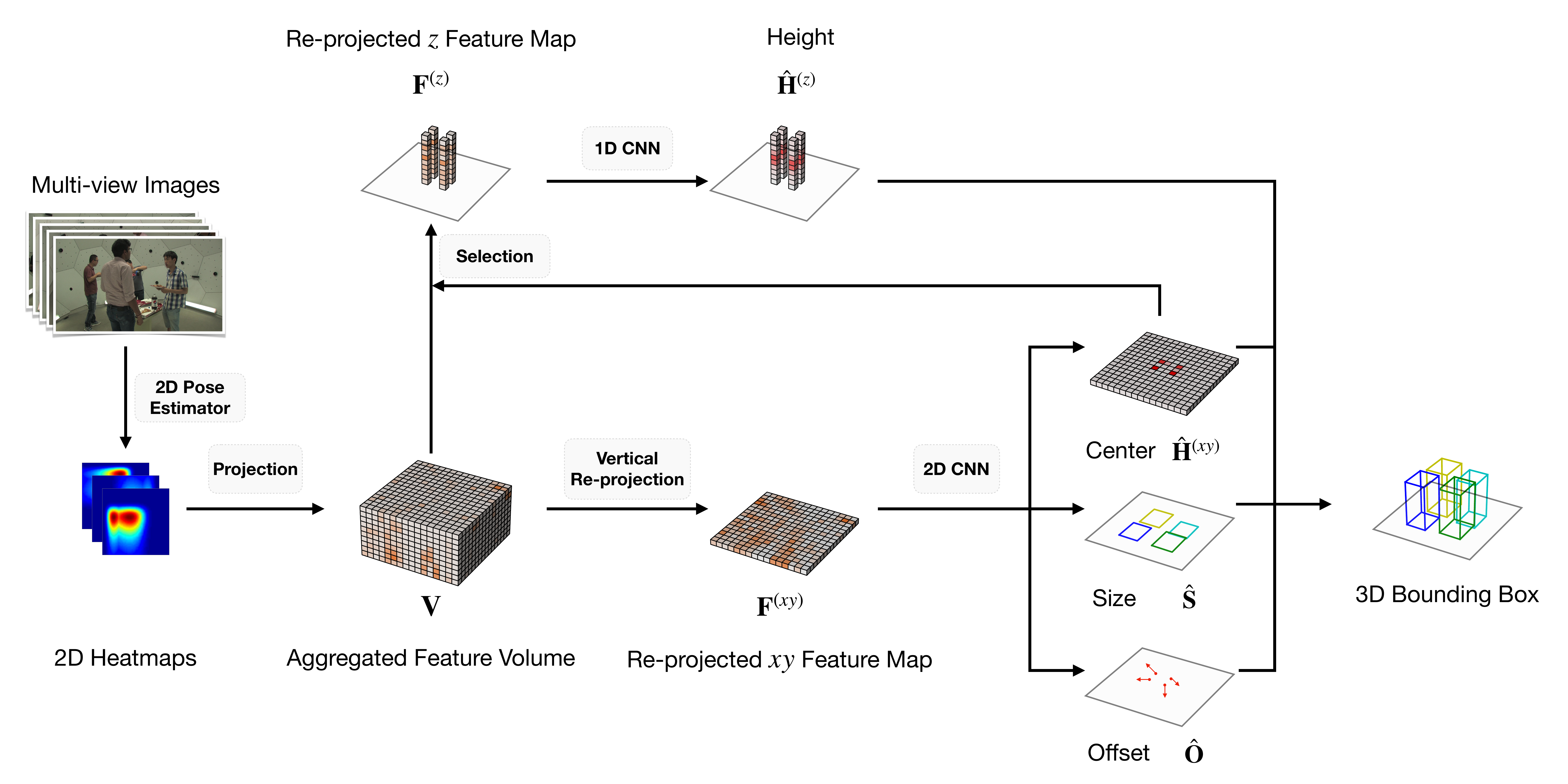}
  \caption{\textbf{Human Detection Networks.} We first construct the feature volume $\mathbf{V}$ from the multi-view images. It is then projected to the $xy$ plane to obtain the feature map $\mathbf{F}^{(xy)}$ (bird's-eye view). A Multi-branch 2D CNN estimates three feature maps encoding each person's center position, bounding box size, and center offset, respectively. We then select the 1D columns feature $\mathbf{F}^{(z)}$ from the positions with high confidence values on $\hat{\mathbf{H}}^{(xy)}$. Then a 1D CNN estimates the heatmap $\hat{\mathbf{H}}^{(z)}$ of the vertical position of the 3D box center. Finally, HDN outputs the combined 3D bounding box.
  }
  \label{fig:hdn}
\end{figure*}

\subsection{Human Detection Networks}
\label{sec:hdn}
We first apply HRNet\cite{sun2019deep} to estimate 2D pose heatmaps from the multiview images, and construct an aggregated feature volume $\mathbf{V} \in \mathbb{R}^{K \times L \times W \times H}$ by back-projecting the heatmaps to the 3D voxel space. Since people are usually on the ground plane and it is less probable that one person is right on top of another, it inspires us to construct a 2D bird's-eye view representation from the feature volume for efficiently detecting people.

\paragraph{Detection in $xy$ Plane}
We re-project the aggregated feature volume to the ground plane ($xy$) by performing max-pooling along the $z$ direction and obtain $\mathbf{F}^{(xy)} \in \mathbb{R}^{K \times L \times W}$. Then we feed $\mathbf{F}^{(xy)}$ to a 2D fully convolutional network to detect the locations of people in the $xy$ plane. The positions of all people in the plane are encoded by a 2D confidence map $ \hat{\mathbf{H}}^{(xy)} \in [0,1]^{L \times W}$ whose value $\hat{\mathbf{H}}^{(xy)}_{i, j}$ represents the likelihood of human presence at the location $(i, j)$. For training supervision, we generate the ground-truth (GT) 2D confidence map $\mathbf{H}^{(xy)}$. Its values are computed by the distance between the GT center point and each grid point using a Gaussian kernel. Specifically, the confidence value of grid point $(i,j)$ is computed by:
\[
   \mathbf{H}^{(xy)}_{i, j} = \max\limits_{1 \le n \le N} \exp \{-\dfrac{(i-\tilde{i_n})^2+(j-\tilde{j_n})^2}{2\sigma^2}\}
\]
where $N$ denotes the number of persons and $(\tilde{i_n}, \tilde{j_n})$ represents the corresponding GT position for person $ n$. We just keep the largest scores in the presence of multiple people. The mean squared error (MSE) loss is computed by:
\begin{equation}
     \mathcal{L}_{2d} = \sum\limits_{i=1}^{L} \sum\limits_{j=1}^{W} \|\mathbf{H}^{(xy)}_{i,j}-\hat{\mathbf{H}}^{(xy)}_{i,j}\|_2
\end{equation}

We further estimate a 2D box size for each person instead of assuming a loose constant size as in the previous work~\cite{voxelpose}. The height of the box is simply set to be $2000$mm. This is critical to isolate the interference of multiple people, especially in crowded scenes. Our model generates a box size embedding at all grid points, denoted as $\hat{\mathbf{S}} \in \mathbb{R}^{2\times L \times W}$. But only those at the locations with large confidences are meaningful. We compute a ground-truth size embedding $\mathbf{S}$ based on box annotations.

During training, we only compute losses on the grid points which are adjacent to the ground-truth box centers. Specifically, for a 2D GT box center $(\Tilde{x}, \Tilde{y})$, we only add supervision on the discretized grid points $(\lfloor \frac{\Tilde{x}}{l}\rfloor, \lfloor\frac{\Tilde{y}}{w}\rfloor)$, where $l$ represents the length of a single voxel and $w$ denotes the width. Let $\mathbf{U}$ denote the set of the neighboring points mentioned above and suppose $N$ is the number of persons in the image. We compute an $L_1$ loss at each center point in $\mathbf{U}$:
\begin{equation}
 \mathcal{L}_{size} = \dfrac{1}{N}\sum\limits_{(i,j) \in \mathbf{U}} \|\mathbf{S}_{i,j}-\mathbf{\hat{S}}_{i,j}\|_1
\end{equation}

In addition, to reduce the quantization error, we estimate the local offset for each root joint on the horizontal plane. Similar to size estimation, the model outputs an offset prediction at each grid point, denoted as $\hat{\mathbf{O}} \in \mathbb{R}^{2\times L \times W}$. We generate a GT offset prediction $\mathbf{O}$ and use an $L_1$ loss on the neighboring points:
\begin{equation}
 \mathcal{L}_{off} = \dfrac{1}{N}\sum\limits_{(i,j) \in \mathbf{U}} \|\mathbf{O}_{i,j}-\hat{\mathbf{O}}_{i,j}\|_1
\end{equation}

Inspired by \cite{zhou2019objects}, we use a simple network structure with three parallel branches to estimate the heatmap, offset and size respectively. As shown in Fig.~\ref{fig:hdn}, the 2D bird's-eye features are passed through a fully-convolutional backbone network and then fed into three separate branches with identical designs, which consist of a 3 $\times$ 3 convolution, ReLU, and another 1 $\times$ 1 convolution. 

\paragraph{Detection in $z$ Axis}
The remaining task is to estimate the center height for each proposal. Firstly, we obtain the proposals with $P$ largest confidences on the 2D heatmap $\hat{\mathbf{H}}^{(xy)}$ after applying non-maximum suppression (NMS). We set $P=10$ in our experiments. Subsequently, we extract the corresponding 1D ``columns'' for each proposal from the aggregated feature volume $\mathbf{V}$, denoted as $\mathbf{F}^{(z)}\in \mathbb{R}^{P \times K \times H}$, which is then fed into a 1D fully convolutional network to regress the height. Similar to 2D detection, our model generates 1D heatmap estimation $\hat{\mathbf{H}}^{(z)} \in [0,1]^{P\times H}$, indicating the likelihood of human presence at every possible height. We compute a GT 1D heatmap  $\mathbf{H}^{(z)}$ for each proposal based on its center height using the Gaussian distribution. Likewise, we use an MSE loss here:
\begin{equation}
    \mathcal{L}_{1d} = \dfrac{1}{P}\sum\limits_{p=1}^P \sum\limits_{k=1}^{H} \|\mathbf{H}_{p,k}^{(z)}-\hat{\mathbf{H}}_{p,k}^{(z)}\|_2
\end{equation}

Finally, we select the height with maximum confidence and by combining it with the 2D box center, offset and size, we can obtain the 3D bounding box. The overall confidence score for each box is computed by multiplying the scores of the 2D heatmap and 1D outputs. According to the exponential property of the Gaussian function, it can be regarded as an approximate of the 3D Gaussian distribution. We set a threshold for confidence scores to select the valid proposals. To sum up, the overall training objective is as follows:
\begin{equation}
    \mathcal{L}_{HDN} = \mathcal{L}_{2d} + \lambda_{size}\mathcal{L}_{size} + \lambda_{off}\mathcal{L}_{off} + \lambda_{1d}\mathcal{L}_{1d}
\end{equation}
where we set $\lambda_{size}=0.02$, $\lambda_{off}=0.1$ and $\lambda_{1d} = 1$.

\subsection{Joint Localization Networks}
\label{sec:JLN}
\begin{figure*}[t]
  \centering
  \includegraphics[width=\linewidth]{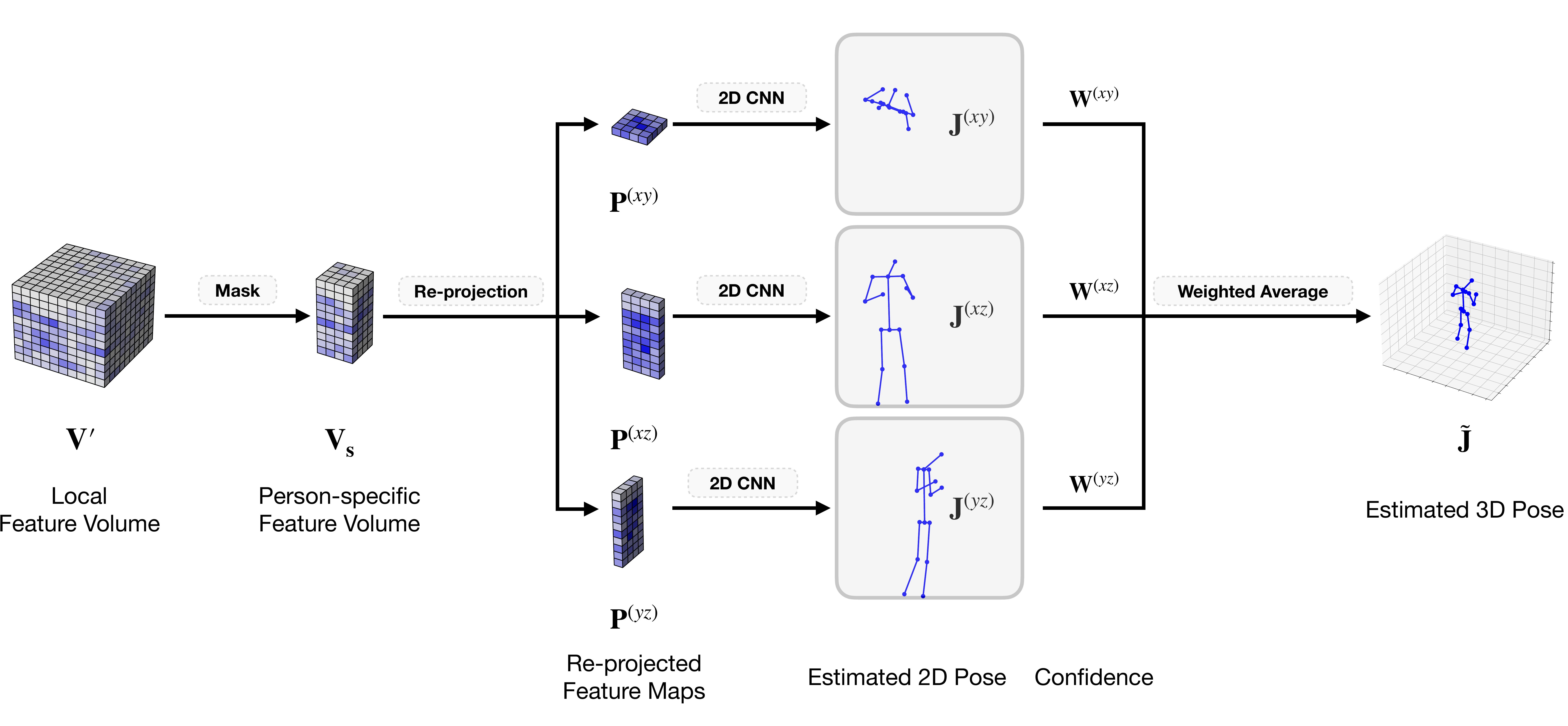}
  \caption{\textbf{Joint Localization Networks.} For each person, we first construct its local feature volume $\mathbf{V'}$. The person-specific feature volume $\mathbf{V_s}$ is obtained by masking $\mathbf{V'}$ with the detected 3D box. We re-project $\mathbf{V_s}$ to three orthogonal coordinate planes to get the 2D feature maps $\mathbf{P}^{(t)}$. A shared 2D pose estimator regresses the joint locations $\mathbf{J}^{(t)}$ for each plane, and a confidence network computes the corresponding weights $\mathbf{W}^{(t)}$. Finally, the 3D pose $\tilde{\mathbf{J}}$ is computed by weighting $\mathbf{J}^{(t)}$ with $\mathbf{W}^{(t)}$ in a pairwise manner. $(t \in \{xy,xz,yz\})$} 
  \label{fig:jln}
\end{figure*}

\paragraph{Person-specific Feature Volume.} With the bounding box of each person, we construct its fine-grained feature volume to predict the final 3D pose. We first crop a smaller feature volume $\mathbf{V'}$ from $\mathbf{V}$ centered at the box center with a fixed size (\ie 2m $\times$ 2m $\times$ 2m). It suffices to cover arbitrary poses and maintains the relative scale of the motion space. The space is then divided into ${L'}\times {W'} \times {H'}$ voxels. Now the key step is to \textbf{zero out} the features outside the estimated bounding box to get the person-specific feature volume $\mathbf{V_{s}}$. This masking mechanism reduces the distraction of other people and enables safe volume re-projection in the following stage. 

\paragraph{Joint Localization.} To reduce the computational cost, we re-project $\mathbf{V_{s}}$ onto three orthogonal 2D planes, \ie the \textit{xy} plane, \textit{xz} plane and \textit{yz} planes in the world coordinate systems. Let $\mathbf{P}^{(xy)} \in \mathbb{R}^{K\times L' \times W'}$, $\mathbf{P}^{(xz)}\in \mathbb{R}^{K\times L' \times H'}$ and $\mathbf{P}^{(yz)} \in \mathbb{R}^{K\times W' \times H'}$ denote the re-projected feature maps corresponding to the three planes, respectively. Again, we use max-pooling for feature projection. 

Subsequently, they are concatenated as a batch and fed to a 2D CNN for joint localization, as shown in Fig.~\ref{fig:jln}. Note that we set the same granularity of voxels on different axes to enable parallel estimation, \ie $L' = W' = H'$. The 2D CNN produces a joint-wise heatmap estimation for each re-projection plane, denoted as $\hat{\mathbf{H}}^{(t)} (t \in \{xy,xz,yz\})$ in the same shape of $\mathbf{P}^{(t)}$. To reduce the quantization error, we compute the center of mass of $\hat{\mathbf{H}}^{(t)}$ instead of taking the maximum responses. Specifically, the estimated positions $\hat{\mathbf{J}}^{(t)} \in \mathbb{R}^{K\times2} $ are computed by:

\begin{align}
 \label{eqn:eqlabel}
\begin{split}
  \hat{\mathbf{J}}^{(xy)} = \sum\limits_{i=1}^{L} \sum\limits_{j=1}^{W} (i,j) \cdot \hat{\mathbf{H}}^{(xy)}_{i,j}
\\
  \hat{\mathbf{J}}^{(xz)} = \sum\limits_{i=1}^{L} \sum\limits_{k=1}^{H} (i,k) \cdot \hat{\mathbf{H}}^{(xz)}_{i,k} 
  \\
  \hat{\mathbf{J}}^{(yz)} = \sum\limits_{j=1}^{W} \sum\limits_{k=1}^{H} (j,k) \cdot \hat{\mathbf{H}}^{(yz)}_{j,k}
 \end{split}
 \end{align}

We supervise the estimations with the ground-truth 2D location $\mathbf{J}^{(t)} \in \mathbb{R}^{K\times 2}$ on each plane. An $L_1$ loss is computed by:
\begin{equation}
 \mathcal{L}_{hm} = \sum\limits_{t}\sum\limits_{k=1}^K\|\mathbf{J}^{(t)}_{k} - \hat{\mathbf{J}}^{(t)}_{k}\|_1
\end{equation}

\paragraph{Adaptive Weighted Fusion.} The quality of $\mathbf{P}^{(t)}$ and the difficulty of pose estimation naturally vary with the re-projection plane and human pose, thus we hope the model could learn to discriminate and balance the estimations from different planes automatically. To achieve this, we introduce a lightweight confidence regression network. We assume that the pattern of $\hat{\mathbf{H}}^{(t)}$ could reflect the quality of 2D pose estimation. Therefore, the estimated heatmaps $\hat{\mathbf{H}}^{(t)}$ are fed into a shared confidence regression network. Inspired by \cite{zhang2021adafuse}, we adopt a simple design for the confidence regression network, consisting of a convolutional layer, a global average pooling layer and one fully-connected layer. 

The network generates joint-wise fusion weight for each plane, denoted as $\mathbf{W}^{(t)} \in \mathbb{R}^{K}$. We then use the Softmax function for normalization in a pair-wise manner and obtain the final 3D prediction $\tilde{\mathbf{{J}}} \in \mathbb{R}^{K \times 3}$. Specifically, for the joint $k$, the final estimations can be computed by: 
\begin{align}
\begin{split}
 \mathbf{\tilde{J}}_{k, 1} = \text{softmax}(\mathbf{W}^{(xy)}_k, \mathbf{W}^{(xz)}_k) \cdot (\hat{\mathbf{J}}^{(xy)}_{k, 1}, \hat{\mathbf{J}}^{(xz)}_{k, 1})
\\
 \mathbf{\tilde{J}}_{k, 2} = \text{softmax}(\mathbf{W}^{(xy)}_k, \mathbf{W}^{(yz)}_k) \cdot (\hat{\mathbf{J}}^{(xy)}_{k, 2}, \hat{\mathbf{J}}^{(yz)}_{k, 1})
 \\
 \mathbf{\tilde{J}}_{k, 3} = \text{softmax}(\mathbf{W}^{(xz)}_k, \mathbf{W}^{(yz)}_k) \cdot (\hat{\mathbf{J}}^{(xz)}_{k, 2}, \hat{\mathbf{J}}^{(yz)}_{k, 2})
\end{split}
\end{align}
where $\hat{\mathbf{J}}^{(xy)}_{k, 1}$ denotes taking the first component of the 2D estimated coordinates of $\hat{\mathbf{J}}^{(xy)}_k$, namely the component on the \textit{x}-axis, and the other notations have similar interpretations. Let $\mathbf{J}$ denote the GT 3D pose, we use an $L_1$ loss to train the confidence regression network:
\begin{equation}
 \mathcal{L}_{conf}  = \sum\limits_{k=1}^K \|\mathbf{J}_k -\mathbf{\tilde{J}}_k\|_1
\end{equation}

Now we get the overall training objective of JLN as follows. In our experiments, we set $\lambda_{conf} = 1$.

\begin{equation}
 \mathcal{L}_{JLN} = \mathcal{L}_{hm} + \lambda_{conf}\mathcal{L}_{conf} 
\end{equation}

\section{Experiments}
\label{sec:experiments}

\subsection{Setup}

\noindent{\textbf{Datasets.}} The Shelf\cite{Belagiannis_2014_CVPR} dataset captures four people disassembling a shelf using five cameras. We select the frames of test set following previous works \cite{voxelpose, Lin_2021_CVPR}. The Campus\cite{Belagiannis_2014_CVPR} dataset captures multiple people interacting with each other in an outdoor environment shot by three cameras. The CMU Panoptic\cite{Joo_2015_ICCV} dataset captures multiple people engaging in social activities. We use the same training and testing sequences captured by five HD cameras as in \cite{voxelpose, Lin_2021_CVPR}.\newline

\noindent{\textbf{Training Strategies.}} Due to incomplete annotations of Shelf and Campus, we use synthetic 3D poses to train the model for the two datasets, following ~\cite{voxelpose, Lin_2021_CVPR}. For the Panoptic dataset, we first finetune the 2D heatmap estimation network. Then we fix the 2D network and train the 3D networks following ~\cite{voxelpose}. \newline

\noindent{\textbf{Evaluation Metrics.}} Following the common practice, we compute the Percentage of Correct Parts (PCP3D) metric on Shelf and Campus. Specifically, we pair each GT pose with the closest estimation and calculate the percentage of correct parts. For the Panoptic dataset, we adopt the Average Precision (AP$_K$) and Mean Per Joint Position Error (MPJPE) as metrics, which reflect the quality of multi-person 3D pose estimation more comprehensively. In addition, we measure the inference time and frame per second (FPS) on the Panoptic dataset.

\begin{figure*}[h]
  \centering
  \includegraphics[width=0.9\linewidth]{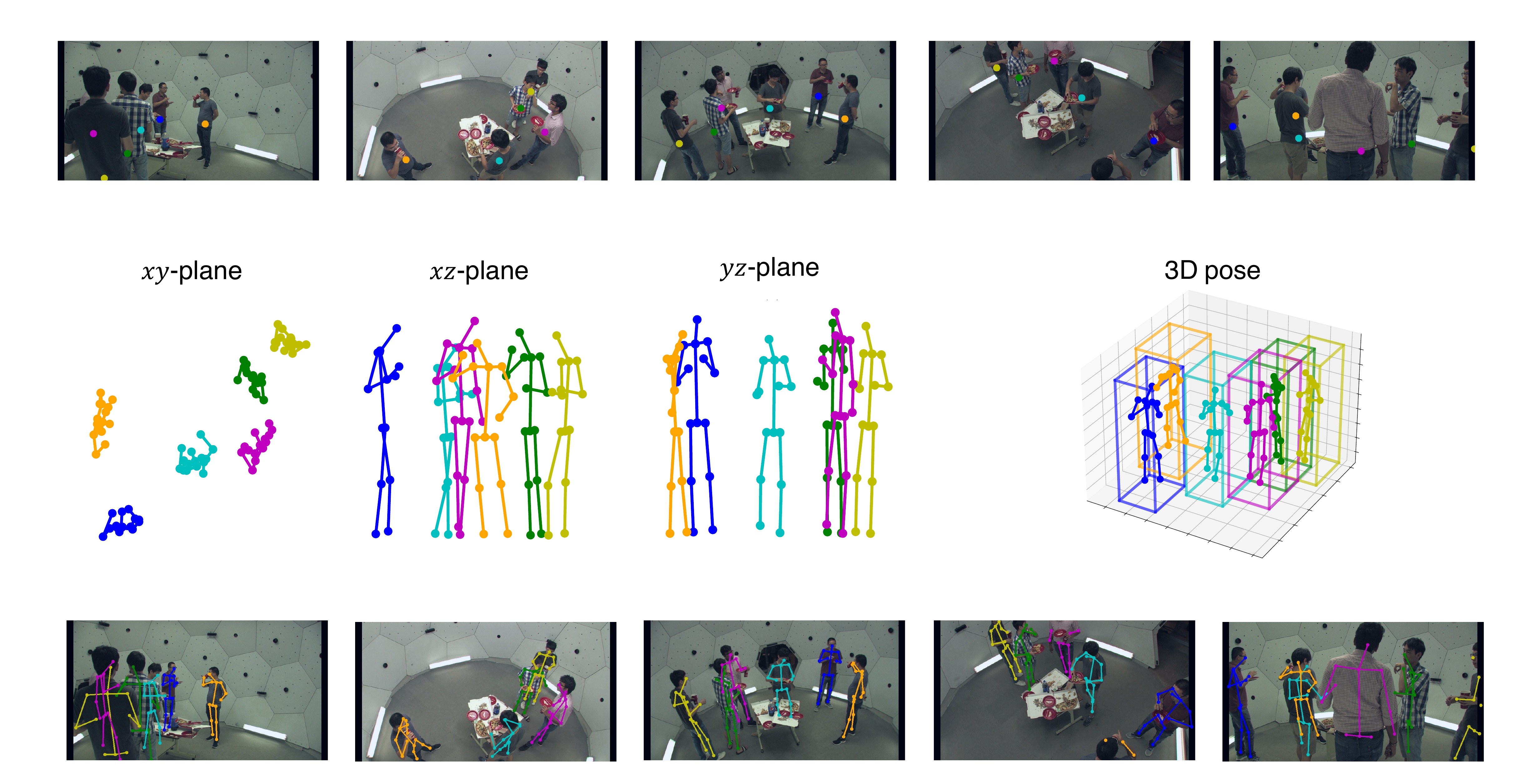}
  \caption{\textbf{Qualitative Results on the CMU Panoptic Dataset.} The first row illustrates the estimated root joints in HDN. The second row shows the estimated 2D poses on the three orthogonal re-projection planes and the fused 3D pose in JLN. The last row shows the 2D back-projection of the estimated 3D pose to each camera view.}
  \label{fig:quality}
\end{figure*}

\begin{table}[b]
\centering
\caption{\textbf{Quantitative Evaluation of HDN.} We measure the mean center error, precision and recall rate to evaluate the quality of human center detection and offset regression. The IoU score is computed between the estimated horizontal bounding box and GT, which additionally reflects the precision of bounding box size estimation.\newline} 
\label{tab:hdn}
\setlength{\tabcolsep}{3mm}
\scalebox{0.9}{
\begin{tabular}{c|c|c|c}
\toprule
Mean Center Error (mm) & Precision & Recall & IoU \\
\midrule
53.73 & 0.9982 & 0.9985 & 0.757 \\
\toprule
\end{tabular}
}
\end{table}

\subsection{Evaluation and Comparison}
\subsubsection{Evaluation of HDN.}

We first evaluate the performance of the Human Detection Networks qualitatively. As Fig.\ref{fig:quality} shows, our model is able to detect the human centers and estimate the 3D bounding boxes as intended, despite the fact that severe occlusion occurs in all views. Accurate 3D bounding boxes help to isolate the persons for the fine-grained joint localization. In addition, we quantitatively measure the performance of HDN in terms of both center position and bounding box. As Tab.\ref{tab:hdn} shows, our HDN localizes the root joint well, and the regressed bounding boxes overlap with GT mostly. The mean center error is larger than the MPJPE of JLN because JLN involves detailed pose estimation on a finer voxel granularity. Still, the center precision suffices to provide a reasonable 3D bounding box for joint localization. \newline

\begin{table*}[h]
    \centering
    \caption{\textbf{Comparison with SOTA on Campus and Shelf.} We compute the PCP3D (Percentage of
Correct Parts) metrics following previous work. A part is considered correct if its distance with GT is at most half of the limb length.\newline}
\setlength{\tabcolsep}{3mm}
\resizebox{\columnwidth}{!}{
\begin{tabular}{r|cccc|cccc}
\toprule
 & \multicolumn{4}{c|}{Shelf} & \multicolumn{4}{c}{Campus} \\
Method & Actor1 & Actor2 & Actor3 & Average & Actor1 & Actor2 & Actor3 & Average\\
\midrule
\midrule

Belagiannis \etal\cite{Belagiannis_2014_CVPR} & 66.1 & 65.0 & 83.2 & 71.4 & 82.0 & 72.4 & 73.7 & 75.8\\
Belagiannis \etal\cite{belagiannis2014multiple} & 75.0 & 67.0 & 86.0 & 76.0 & 83.0 & 73.0 & 78.0 & 78.0\\
Belagiannis \etal\cite{belagiannis20153d} & 75.3 & 69.7 & 87.6 & 77.5 & 93.5 & 75.7 & 84.4 & 84.5\\
Ershadi-Nasab \etal\cite{ershadi2018multiple} & 93.3 & 75.9 & 94.8 & 88.0 & 94.2 & 92.9 & 84.6 & 90.6\\
Dong \etal\cite{dong2019fast} & 98.8 & 94.1 & 97.8 & 96.9 & 97.6 & 93.3 & 98.0 & 96.3\\
Huang \etal\cite{DBLP:conf/eccv/HuangJLZTDFX20} & 98.8 & 96.2 & 97.2 & 97.4 & 98.0 & 94.8 & 97.4 & 96.7\\
Tu \etal\cite{voxelpose} & 99.3 & 94.1 & 97.6 & 97.0 & 97.6 & 93.8 & 98.8 & 96.7\\
Lin \etal\cite{Lin_2021_CVPR} & 99.3 & 96.5 & 98.0 & 97.9 & 98.4 & 93.7 & 99.0 & 97.0\\
Wang \etal\cite{wang2021mvp} & 99.3 & 95.1 & 97.8 & 97.4 & 98.2 & 94.1 & 97.4 & 96.6\\
Ours & 99.4 &96.0	&97.5	&97.6 & 96.5 & 94.1 & 97.9 & 96.2\\
\toprule
\end{tabular}}
    
    \label{tab:acc}
\end{table*}

\begin{table}[h]
    \centering
    \caption{\textbf{Comparison with SOTA on Panoptic.} For efficiency metrics, We measure the average per-sample inference time on Panoptics test set (5 camera views, 3.41 person per frame). The measurement is done on a Linux machine with GPU GeForce RTX 2080 Ti and CPU Intel(R) Xeon(R) CPU E5-2699A v4 @ 2.40GHz. Batch size is set to be $1$ for all methods.\newline}
    \setlength{\tabcolsep}{3mm}
\resizebox{0.85\columnwidth}{!}{
    \begin{tabular}{c|ccccccc}
\toprule
Method & AP$_{25}$ & AP$_{50}$ & AP$_{100}$ & AP$_{150}$ & MPJPE & Time & FPS\\
\midrule
VoxelPose\cite{voxelpose} & 83.59& 98.33 &99.76 & 99.91& 17.68mm & 316.0ms & 3.2\\
PlaneSweepPose \cite{Lin_2021_CVPR} & 92.12& 98.96 &99.81 & 99.84& 16.75mm & 234.3ms & 4.3\\
MvP \cite{wang2021mvp} & 92.28& 96.60 &97.45 & 97.69& 15.76mm & 278.8ms & 3.6\\
Ours & 85.22 & 98.08 & 99.32 & 99.48 &18.26mm & 32.2ms & 31.1\\

\toprule
\end{tabular}}
    
    \label{tab:panoptic}
\end{table}
\subsubsection{Evaluation of JLN.} We compare the 3D pose estimation performance with the state-of-the-art (SOTA) multi-view multi-person 3D pose estimation methods on Shelf and Campus~\cite{Belagiannis_2014_CVPR}. While the proposed method is primarily optimized for inference efficiency and makes several approximations, it performs competitively with SOTA as shown in Tab.\ref{tab:acc}. On the Shelf dataset, it outperforms the SOTA volumetric approach VoxelPose~\cite{voxelpose} which features fully 3D convolutional architecture. We also train and test on the Panoptic~\cite{Joo_2015_ICCV} dataset following the most recent works~\cite{voxelpose, Lin_2021_CVPR}. As shown in Tab.~\ref{tab:panoptic}, our method receives an extra per-joint error of about 2mm. We argue that the error margin is within an acceptable range given the speed-accuracy trade-off in real-time applications. \newline

\subsubsection{Efficiency.} We first compare the inference speed of our method to the SOTA methods, and then conduct an in-depth efficiency analysis. The speed results of other methods are obtained using their official codes on the same hardware as ours. For a fair comparison, we set the batch size to be one for all methods during inference following \cite{wang2021mvp} to simulate the real-time use case where data arrives frame by frame. The batch size of PlaneSweepPose\cite{Lin_2021_CVPR} was set to be $64$ in the original paper so their reported speed is different from the one reported in this paper. For all the methods, the off-the-shelf 2D pose estimator time is not measured following\cite{wang2021mvp, Lin_2021_CVPR}. The results on the Panoptic dataset are shown in Tab.~\ref{tab:panoptic}. Our approach shows a considerable advantage in terms of inference speed and supports real-time inference.

\begin{figure}[t]
  \centering
  \includegraphics[width=0.9\linewidth]{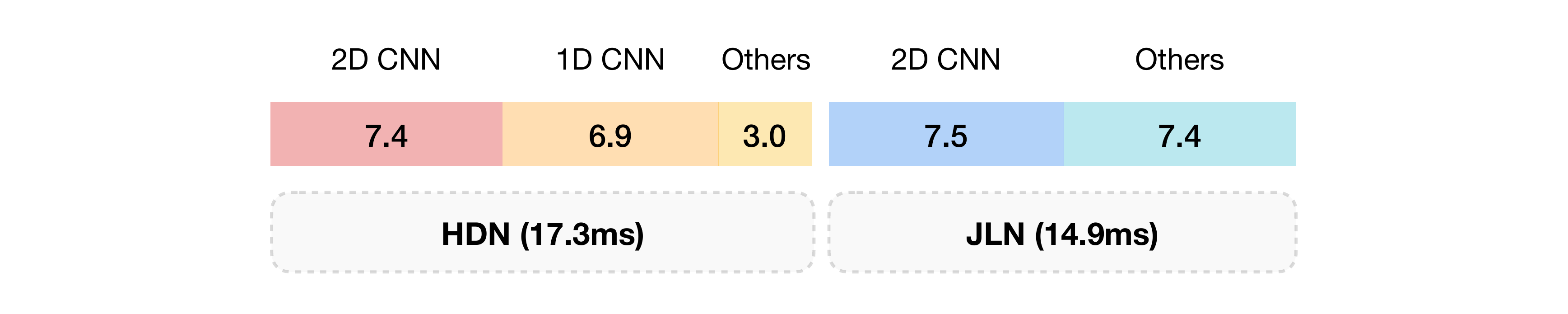}
  \caption{\textbf{Time Cost Visualization.} We compute the average inference time cost for each module on the Panoptic test set in milliseconds. It takes $32.2$ms in total.}
  \label{fig:time}
\end{figure}

The inference time broken down per module is shown in Fig.~\ref{fig:time}. The ``others'' parts mainly consist of data preparation and feature volume construction. For HDN, the time cost is independent of the number of cameras and persons. For JLN, the theoretical computation complexity is linear to the number of persons. In practice, feature maps of different persons are concatenated as a batch and inferred in a single feedforward. As we only use the re-projected 2D feature maps, the batch size could be large enough to cover very crowded scenes. In general, the time cost of our method is mainly determined by voxel granularity. By using $2\times$ coarser voxel, its computation complexity could be reduced to $\frac{1}{4}$. The voxel granularity selection serves as a trade-off between speed and accuracy.

Finally, we analyze the scalability of our method and compare it with the existing methods. Consider applying the algorithms to a challenging scenario that is much larger and more crowded than the current datasets~\cite{Joo_2015_ICCV, Belagiannis_2014_CVPR}. In order to retain a reasonable coverage, the number of cameras needs to grow proportionally~\cite{Zhang15acamera, xu2020hitmac}. VoxelPose~\cite{voxelpose} uses massive 3D convolution operations that are computation-intensive, and its efficiency disadvantage would be enlarged when scaling. PlaneSweepPose~\cite{Lin_2021_CVPR} needs to enumerate the depth planes for every pair of camera views and persons. As a result, the computation complexity increases in polynomials regarding the number of cameras and persons. For example, simply shifting from Campus (3 persons, 3 cameras) to Shelf (4 persons, 5 cameras) slows PlaneSweepPose by $2.6\times$ according to ~\cite{Lin_2021_CVPR} ($1.3 \times$ for our method). MvP~\cite{wang2021mvp} uses projective attention to integrate the multi-view information, and its time cost also grows quadratically as camera number increases. As previously analyzed, our method does not involve explicit view-person association, and its speed is mainly affected by the granularity of space division. We argue that the above characteristics make our method more scalable to large, crowded scenes than the previous methods. We deployed our model to a basketball court and a retail store where the space size is $16$m$\times16$m with $12$ cameras and $10$ people. Our inference time increases by $28.8\%$ compared to that on Panoptic ($8\text{m}\times8\text{m}$, $5$ cameras, $3.4$ persons).

\subsection{Ablation Study}

We train some ablated models to study the impact of the individual factors. All the ablation experiments are evaluated on CMU Panoptic~\cite{Joo_2015_ICCV}, and the results are shown in Tab. \ref{tab:ablation}.\newline

\noindent\textbf{Feature Masking.}
In (b), we remove the masking step and directly use the local feature volume $\mathbf{V}'$ in JLN. This is equivalent to using a fixed bounding box size as ~\cite{voxelpose}. The degraded performance indicates that the masking mechanism indeed reduces the ambiguity and helps joint localization.\newline

\noindent\textbf{Adaptive Weighted Fusion.}
In (c), we simply take the mean of the estimated coordinates from different planes to compute the final result. The performance gap suggests that the learned confidence weights emphasize the more reliable estimations as intended.\newline

\noindent\textbf{Number of Cameras.} In (d)-(e), we compare the performance under different camera numbers. The accuracy drops with fewer camera views as the feature volume coverage is weakened. \newline

\noindent\textbf{Granularity of Voxels.}
We study the impact of voxel granularity on both efficiency and accuracy. Tab. \ref{tab:ablation_voxel}. shows models trained with different JLN voxel sizes. By reducing the number of voxels (effectively increasing the individual voxel size), the error increases slightly while the inference efficiency improves. It inspires us to balance the trade-off between speed and accuracy in real usage.

\begin{table}[t]
    \centering
    \setlength{\abovecaptionskip}{8pt}
    \caption{\textbf{Ablation Study Results.} Our full approach is (a). From (b) to (e), we study the effect of volume feature masking, weighted fusion and camera views respectively.}
    \resizebox{0.7\columnwidth}{!}{
    \begin{tabular}{c|cccccccccc}
\toprule
Method & \#Views &  Mask & Weighted & AP$_{25}$ & AP$_{50}$ & AP$_{100}$ &  AP$_{150}$ & MPJPE\\
\midrule
(a) & 5 &  \checkmark &\checkmark&85.22 & 98.08 & 99.32 & 99.48 &18.26mm\\
(b) & 5 &   &\checkmark&72.05 & 96.75	&99.10 &	99.39&	21.07mm\\
(c) & 5 &  \checkmark & & 77.23 & 97.61& 99.18& 99.48& 20.11mm\\
\midrule
(d) & 4 &  \checkmark &\checkmark&73.95 & 97.02	&99.21 &99.35&	21.12mm \\
(e) & 3 &  \checkmark &\checkmark&53.68 & 91.89	&97.40 &98.30&	26.13mm \\
\toprule
\end{tabular}}
\label{tab:ablation}
\end{table}

\begin{table}[t]
    \centering
    \setlength{\abovecaptionskip}{8pt}
    \caption{\textbf{Influence of Voxel Granularity.} We additionally report the MACs (Multiply–Accumulate Operations) and number of parameters of the networks.}
    \setlength{\tabcolsep}{3mm}
\resizebox{0.9\columnwidth}{!}{
    \begin{tabular}{c|ccccccc}
\toprule
JLN Voxels & AP$_{25}$ & AP$_{50}$ & AP$_{100}$ & AP$_{150}$ & MPJPE & MACs & Parameters\\
\midrule
$64\times64\times64$ & 85.22 & 98.08 & 99.32 & 99.48 &18.26mm & 8.670G & 1.236M\\
$48\times48\times48$ & 78.76& 97.14 &98.99 & 99.14& 19.66mm & 4.877G & 1.210M\\
$32\times32\times32$ & 73.20& 97.37 &98.93 & 99.08& 20.47mm & 2.167G & 1.190M\\

\toprule
\end{tabular}}
    
\label{tab:ablation_voxel}
\end{table}

\clearpage
\section{Conclusion}
\label{sec:conclusion}

In this paper, we present a novel method for 3D human pose estimation from multi-view images. Our pipeline uniquely integrates the feature volume re-projection to both human detection and joint localization, which substitutes the computation-intensive 3D convolutions. Experiment results prove the effectiveness of the proposed HDN and JLN. The accelerated inference demonstrates the potential of our method in real-time applications, especially for large scenes.

\section*{Acknowledgement}
This work was supported in part by MOST-2018AAA0102004 and NSFC-62061136001.

\clearpage
\bibliographystyle{splncs04}
\bibliography{egbib}

\clearpage
\section*{Appendix}
\appendix
\section{Implementation details}
\label{sec:1}
\subsection{Human Detection Networks}
Following \cite{voxelpose}, we discretize the overall motion space into $L\times W \times H$ voxels. In our experiments, we set $L= W=80$ and $H=20$. 

Inspired by \cite{voxelpose}, we adopt a similar Encoder-Decoder architecture in the Human Detection Networks. The key difference is that we replace all expensive 3D convolutions with 2D and 1D convolutions. The basic components of our fully-convolutional networks include vanilla convolutional block and residual convolution block. The former is comprised of one convolutional layer, one batch-norm layer and ReLU while the latter consists of two consecutive basic convolutional blocks with residual connection. At the initial stage, the feature volume is fed into a 7 $\times$ 7 convolutional layer. In the subsequent Encoder structure, the feature representation is downsampled through three 3 $\times$ 3 residual convolutional blocks with maxpooling. The Decoder adopts a symmetric design, but with deconvolution operations. Finally, the network generates the results through a 1 $\times$ 1 convolutional layer. Following \cite{zhou2019objects}, the outputs of 2D networks are fed into three branches to estimate the feature map, the local offset and the size of the bounding box respectively. They share an identical design, which consists of a 3 $\times$ 3 convolution, ReLU and another 1 $\times$ 1 convolution. 

The 1D convolutional network shares the same architecture with its 2D counterpart except for two aspects: 1) all convolutional operations are replaced with 1D convolutions 2) we just maintain the branch for estimating feature maps along the $z$ axis.

\subsection{Joint Localization Networks}
The architecture of Joint Localization Networks is essentially the same as one 2D CNN branch of HDN. The outputs of 2D estimators are further fed into a shared confidence network, which consists of one convolutional layer, one global average pooling layer plus a fully-connected layer.

\subsection{Training}
We train HDN and JLN jointly to convergence.
On the CMU Panoptic dataset, our model is trained $10$ epochs with batch size $8$. On the Shelf and Campus datasets, we train our model for 30 epochs with the same batch size. The learning rate is set to be $\alpha=0.0001$ using Adam \cite{kingma2014adam} optimizer. The parameters above are empirically determined.

In the bounding box regression branch of HDN, we add a safety margin $\delta = 200$mm to GT, as missing information of body joints will be fatal to the subsequent prediction.

\section{Experimental Details}
\label{sec:2}
\subsection{Dataset}
\subsubsection{CMU Panoptic\cite{Joo_2015_ICCV}} This dataset captures multiple people engaging in social activities in an indoor setting. It contains massive sequences in various scenarios. We use the sequences captured by five HD cameras (3, 6, 12, 13, 23). The training and testing split is identical with \cite{voxelpose, Lin_2021_CVPR}. 

\subsubsection{Shelf\cite{Belagiannis_2014_CVPR}} This dataset captures four people disassembling a shelf using five cameras. We follow previous works \cite{voxelpose, Lin_2021_CVPR, dong2019fast, DBLP:conf/eccv/HuangJLZTDFX20} in evaluating only three of the four persons on the test set frames 300-600 since one person is severely occluded. Due to the lack of complete annotations of ground-truth poses, we train with synthetic heatmaps following previous works \cite{voxelpose, Lin_2021_CVPR, Wu_undated-wk}.

\subsubsection{Campus\cite{Belagiannis_2014_CVPR}} This dataset captures multiple people interacting with each other in an outdoor environment by three cameras. We follow previous works~\cite{voxelpose, Lin_2021_CVPR, dong2019fast, DBLP:conf/eccv/HuangJLZTDFX20} and perform evaluation on the test set frames 350-470, 650-750. Similar to the Shelf dataset, we also conduct training on synthetic heatmaps.

\subsection{Evaluation Metrics}
\subsubsection{PCP} For the Percentage of Correct Parts, we pair each GT pose with the closest estimation and calculate the percentage of correct parts. Specifically, the match is counted as correct if their distance is within a threshold $T$. Following ~\cite{voxelpose, Lin_2021_CVPR, dong2019fast, DBLP:conf/eccv/HuangJLZTDFX20}, we set $T$ to be half of the corresponding limb length. Note that PCP does not penalize false positive results.

\subsubsection{AP$_K$} In order to evaluate the results more comprehensively, we follow ~\cite{voxelpose, Lin_2021_CVPR} to measure the Average Precision (within $K$mm). Specifically, a predicted joint is considered as correct if there is a corresponding GT joint within distance threshold $K$.

\subsubsection{MPJPE} We first pair the nearest GT for each predicted joint, then calculate the corresponding Mean Per Joint Position Error in millimeters.

\section{Additional Results}
\label{sec:3}
\begin{figure*}[b]
\centering
\begin{subfigure}{0.75\textwidth}
    \includegraphics[width=\textwidth]{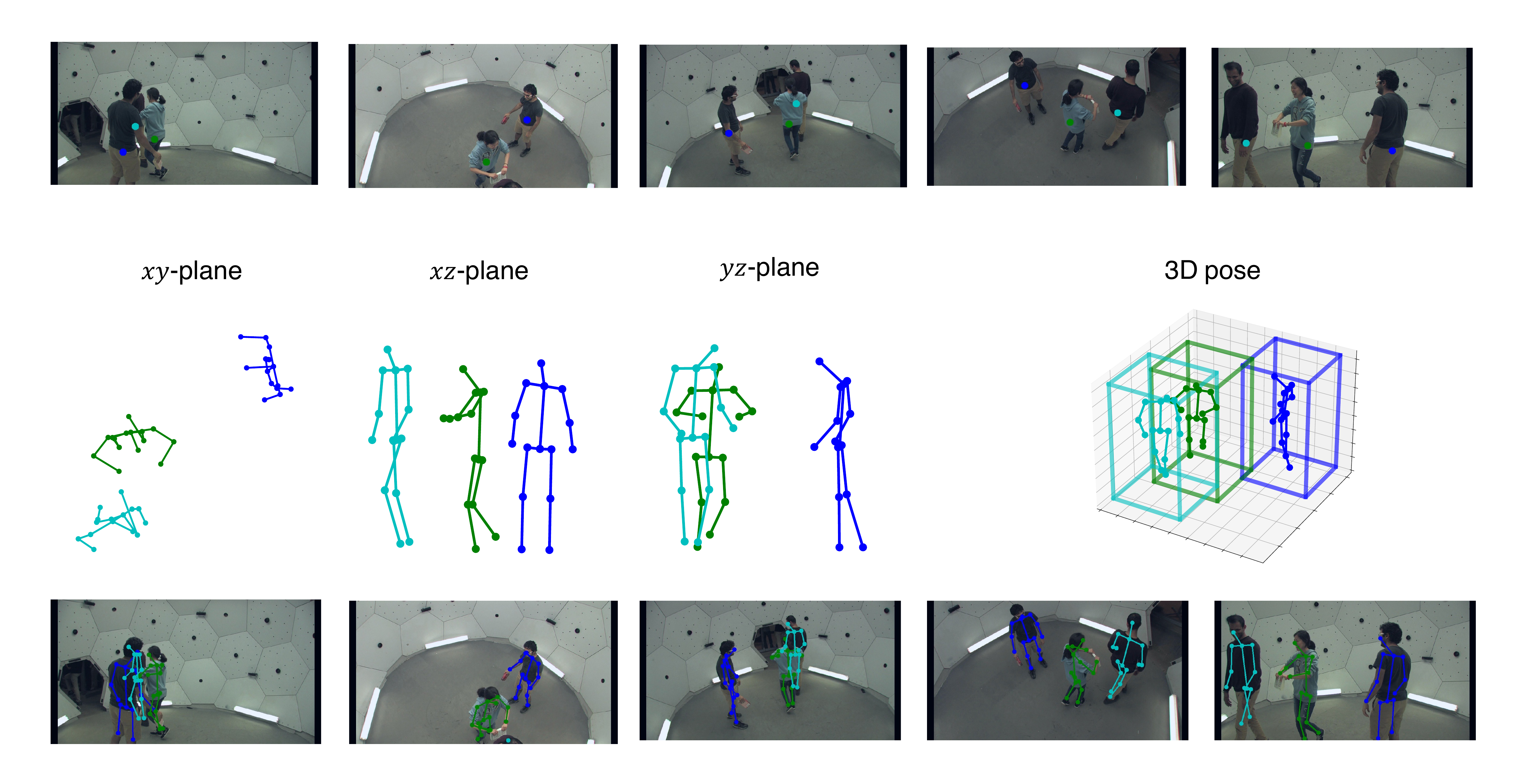}
    \caption{Results on sequence \textit{Haggling}.}
    \label{fig:first}
\end{subfigure}
\hfill
\begin{subfigure}{0.75\textwidth}
    \includegraphics[width=\textwidth]{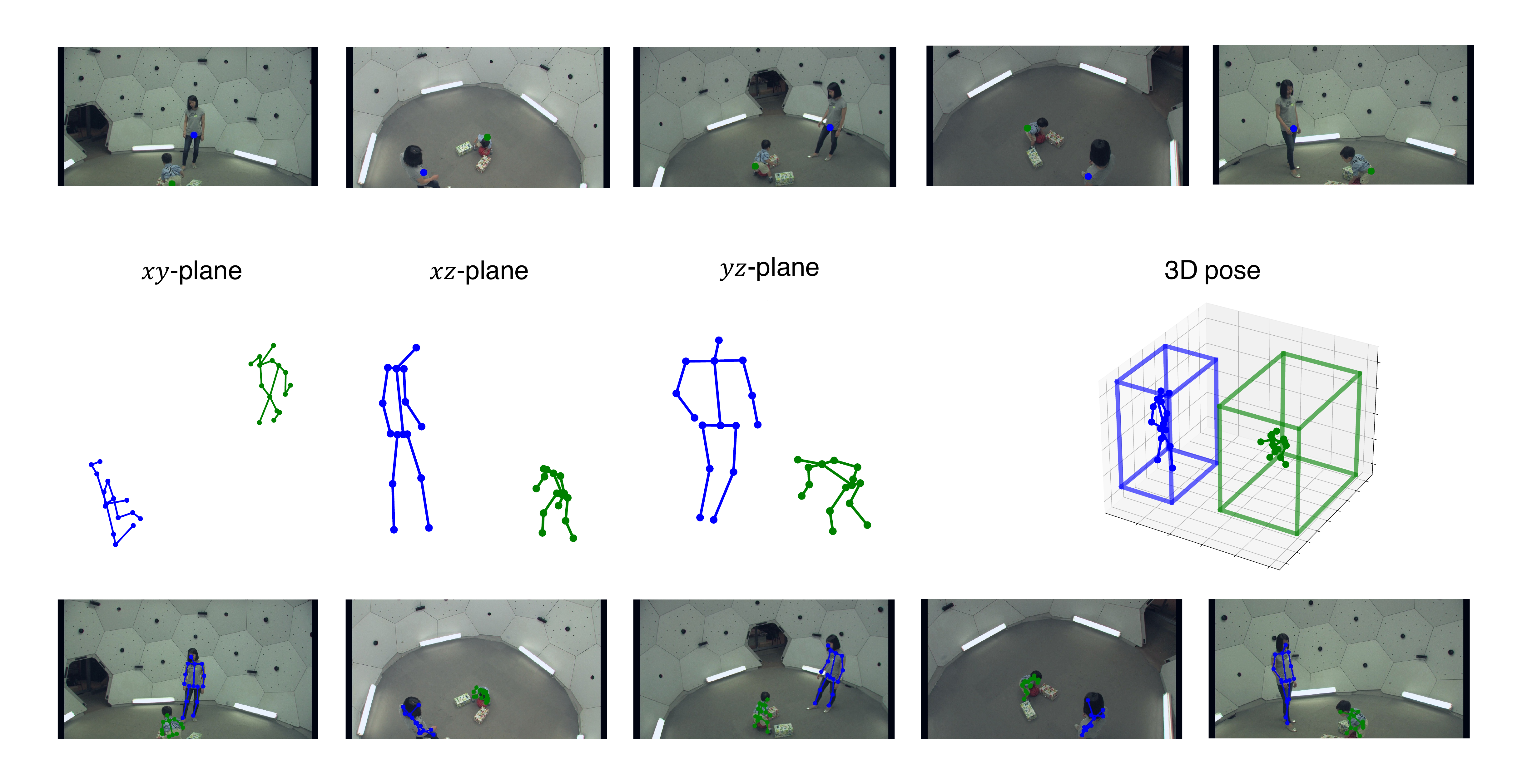}
    \caption{Results on sequence \textit{Ian}.}
    \label{fig:second}
\end{subfigure}
\hfill
\begin{subfigure}{0.75\textwidth}
    \includegraphics[width=\textwidth]{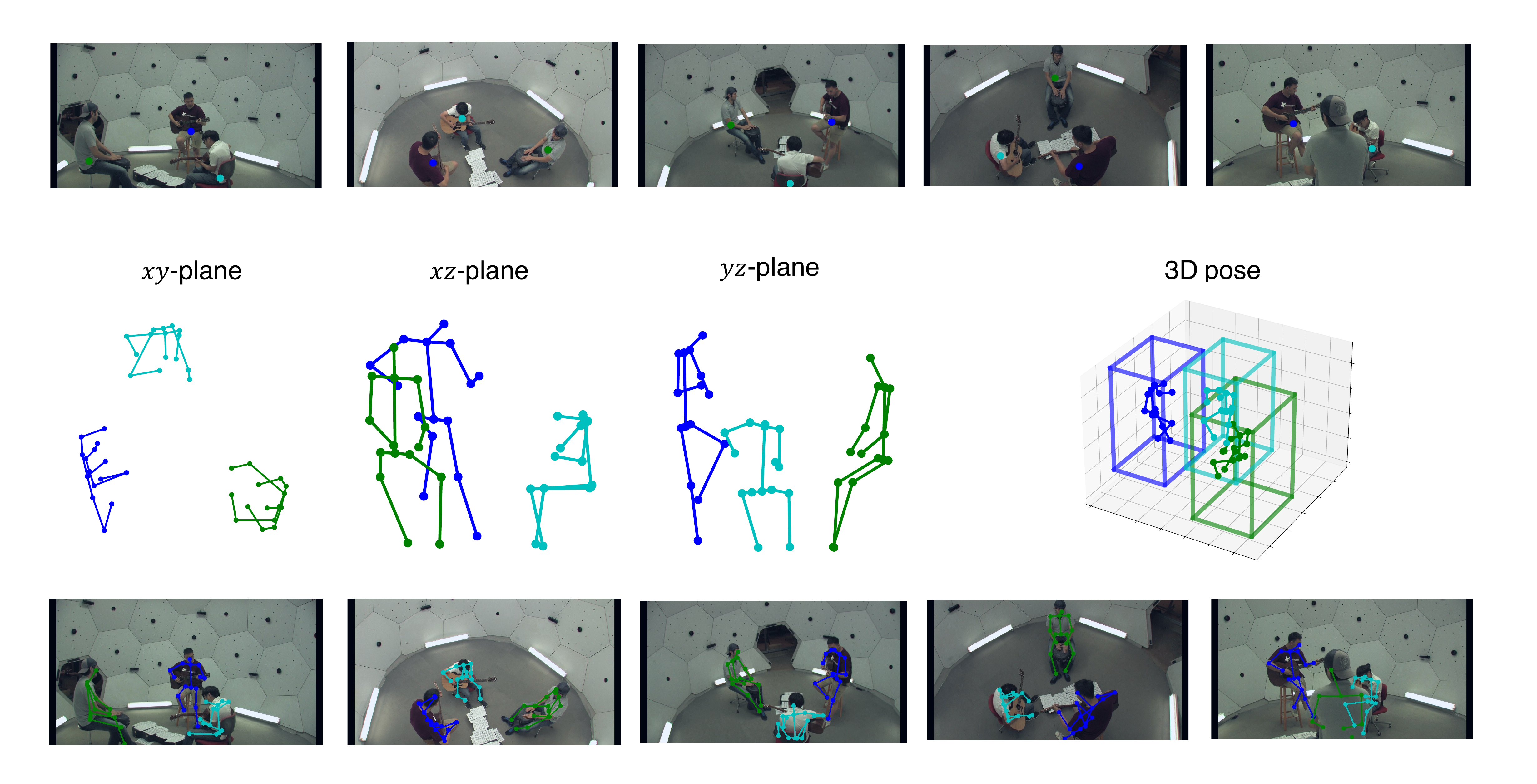}
    \caption{Results on sequence \textit{Band}.}
    \label{fig:third}
\end{subfigure}
        
\caption{\textbf{Additional Results on the CMU Panoptic Dataset.} We present the results on three different action sequences. For each figure, the first row illustrates the estimated root joints in HDN. The second row shows the estimated 2D poses on three orthogonal re-projection planes and the fused 3D pose in JLN. The last row shows the 2D back-projection of the estimated 3D pose to each camera view.}

\label{fig:addtional}
\end{figure*}

We present additional qualitative results in Fig.~\ref{fig:addtional}. Please refer to the attached video for more results.

\clearpage
In addition, we study the influence of the number of persons on inference time. The results are shown in Table.~\ref{tab:time}. The time increase is mainly on the feature construction phase of JLN. 

\begin{table}[h]
\centering
\setlength{\tabcolsep}{1.8mm}
\caption{\textbf{Experiment of scalability.} We measure the average inference time cost of each module in milliseconds (ms) while varying the number of persons present in the synthetic scene.\newline}
\begin{tabular}{c|cccccccccc}
\toprule
Num. & 1 & 2 & 3 & 4 & 5 & 6 & 7 & 8 & 9 & 10\\
\midrule
HDN & 18.27& 17.90 &17.71 & 18.22& 18.28 & 18.50 & 17.37 & 17.86 & 18.30 & 18.45\\
JLN & 13.16& 13.67 &14.22 & 15.03& 16.72 & 18.40 & 20.70 & 21.22 & 24.01 & 25.88\\
Total & 31.43& 31.57 & 31.93 & 33.25& 35.00 & 36.90 & 38.07 & 39.08 & 42.31 & 44.33\\
\toprule
\end{tabular}
\label{tab:time}
\end{table}

\clearpage

\end{document}